%% file: iros2020.tex
\documentclass[letterpaper, 10pt, conference]{ieeeconf}
\IEEEoverridecommandlockouts 
\usepackage{times} 
\usepackage{amsmath} 
\usepackage{amssymb}  
\usepackage{bm}  

\usepackage{array}
\usepackage{epsfig}
\usepackage{etoolbox}
\usepackage{graphics}
\usepackage{makecell}
\usepackage{microtype}
\usepackage{stfloats}
\usepackage[dvipsnames]{xcolor}
\newcommand{\bq}{\mathbf{q}}

\usepackage{tikz}
\usetikzlibrary{%
    decorations.pathreplacing,%
    decorations.pathmorphing%
}

\usepackage[noadjust]{cite} 
\usepackage[font=small]{subcaption}

\usepackage[hidelinks]{hyperref}
\usepackage{paralist}

\usepackage[binary-units=true]{siunitx}

\newcommand{\vq}{\mathbf{q}}
\newcommand{\vqdot}{\mathbf{\dot{q}}}
\newcommand{\vqddot}{\mathbf{\ddot{q}}}
\newcommand{\vx}{\mathbf{x}}

\newcommand{\vu}{\mathbf{u}}
\newcommand{\vv}{\mathbf{v}}
\newcommand{\vf}{\mathbf{f}}
\newcommand{\vr}{\mathbf{r}}

\newcommand{\vg}{\mathbf{G}}
\newcommand{\bM}{\mathbf{M}}
\newcommand{\bC}{\mathbf{C}}
\newcommand{\bS}{\mathbf{S}}
\newcommand{\bJ}{\mathbf{J}}
\newcommand{\vlam}{\pmb{\lambda}}

\RequirePackage{scrlfile}
\PreventPackageFromLoading{subfig}
\usepackage{svg}
\usepackage{hyperref}
\usepackage{xparse}

\usepackage[latin1]{inputenc}

\usepackage{graphicx}
\graphicspath{{figures/}}
\usepackage[export]{adjustbox} 

\usepackage[style=long,nolist]{glossaries}
\glsdisablehyper
\include{glossary}

\usepackage[english]{babel}
\newcommand{\comment}[1]{} 

\title{\LARGE \bf
    Modeling and Control of a Hybrid Wheeled Jumping Robot
}

\author{Traiko Dinev$^1$, Songyan Xin$^{1,3}$, Wolfgang Merkt$^{1,2}$, Vladimir Ivan$^1$, and Sethu Vijayakumar$^1$%
\thanks{$^1$Authors are with the School of Informatics, The University of Edinburgh, Edinburgh, UK.}
\thanks{$^2$Authors are with the Oxford Robotics Institute, University of Oxford, UK.}
\thanks{$^3$Authors are visiting researchers at the Shenzhen Institute for Artificial Intelligence and Robotics for Society (AIRS), CUHK-SZ, China.}
\thanks{Email: {\tt\small traiko.dinev@ed.ac.uk}.}%
}

\begin{document}
\bstctlcite{IEEEexample:BSTcontrol} 

\maketitle
\thispagestyle{empty}
\pagestyle{empty}

\begin{abstract}
    In this paper, we study a wheeled robot with a prismatic extension joint. This allows the robot to build up momentum to perform jumps over obstacles and to swing up to the upright position after the loss of balance. We propose a template model for the class of such two-wheeled jumping robots. This model can be considered as the simplest wheeled-legged system. We provide an analytical derivation of the system dynamics which we
    use inside a model predictive controller (MPC). We study the behavior of the model and demonstrate highly dynamic motions such as swing-up and jumping. 
    Furthermore, these motions are discovered through optimization from first principles. We evaluate the controller on a variety of tasks and uneven terrains in a simulator.
\end{abstract}


\section{Introduction}
Wheels have been used by humanity since the Bronze age. 
Wheeled vehicles have enabled fast and reliable transport of goods across great distances due to their simplicity and efficiency.
However, they require structured terrain such as roads or rails. 
Legged robots on the other hand are capable of navigating unstructured, rough terrain, including jumping over obstacles and across gaps. 
This comes at the cost of a more complex mechanical structure. This complexity makes the robot more expensive to control---both in terms of computation as well as energy expenditure. Wheel-legged robots can combine the best of both designs---the fast motion of a wheeled system with the ability to navigate rugged terrain via legged locomotion.

For model-based control, a model of the system is needed. It should be able to handle the rolling contact of the wheels as well as the discrete contact changes typical for stepping and jumping motions of legged robots. However, instead of adding complexity to a legged robot model by adding the rolling contact, we choose to focus on developing a simplified model with only one prismatic joint to `mimic' the capability of a leg. This provides us with a better tool for modeling the robot behavior than some of the more complex models.



\begin{figure}[t]
    \centering
    \captionsetup{justification=centering}
    \includegraphics[width=\columnwidth]{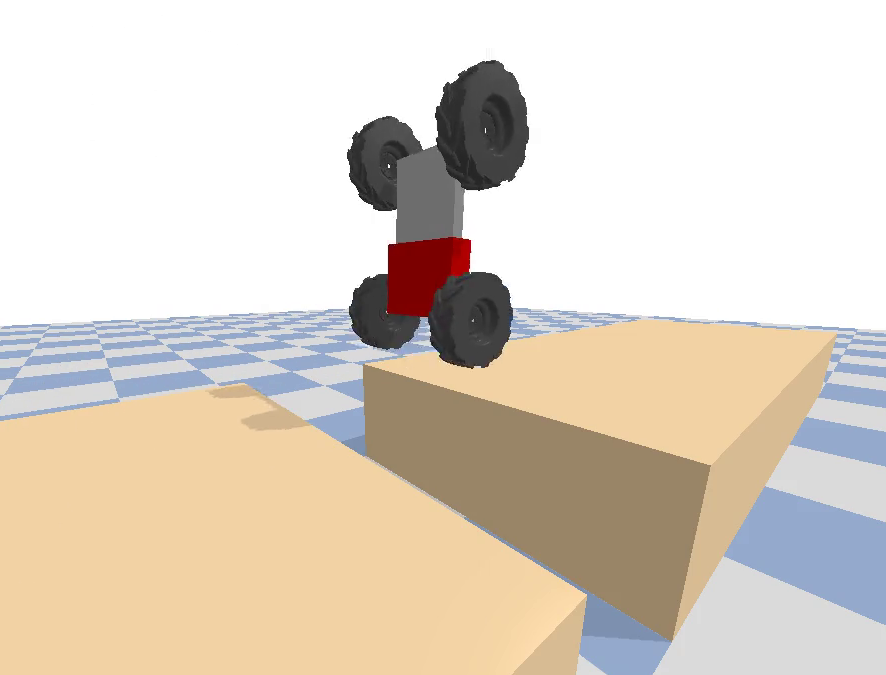}
    \caption{The wheeled jumping robot jumps across a gap. Please find the accompanying video at \\ \url{https://youtu.be/j2sIWL8m2pQ}}
    \label{fig:illustration}
\end{figure}

Most wheel-legged robots (\cite{klemm_ascento:_2019, bjelonic_rolling_2020,bjelonic_keep_2019,geilinger_skaterbots:_2018, giftthaler_efficient_2017}) are high degree of freedom nonlinear systems which do not lend themselves readily to online whole-body trajectory optimization (TO).

Previous work~\cite{giordano_kinematic_2009, giftthaler_efficient_2017, nagano_stable_2015} explored kinematic motion planning where the robot is controlled like a mobile vehicle and the legs act as a suspension. By definition, these techniques do not consider the dynamics of the system.

The authors of~\cite{klemm_ascento:_2019} use a linear quadratic regulator (LQR) for balancing the Ascento robot. They linearize the nonlinear system dynamics around the fixed point at the upright configuration. This provides an efficient approximation but it limits the range of motion around the fixed point \cite[Chapter~3]{tedrake_underactuated_2018}.

By contrast, ANYmal~\cite{bjelonic_rolling_2020} uses a centroidal dynamics model~\cite{orin_centroidal_2013} in a two-stage controller: One for the center of mass (CoM) and one for the wheels. This approach, therefore, fails to exploit potential combined synergies of the wheel and body dynamics.

Skaterbots~\cite{geilinger_skaterbots:_2018} solve the full optimization. However, due to the complexity of the resulting nonlinear program, the constraints are expressed as cost terms and optimized using Newton's method. Compared to Skaterbots, we are able to solve the nonlinear problem due to the simple template model. Additionally, here, we use a Model Predictive Controller scheme, compared to the PD control used in Skaterbots.

Finally, the authors in \cite{de_viragh_trajectory_2019} propose a Quadratic Programming (QP)-based approach to wheel-legged locomotion. In order to make the problem tractable, the $z$-component of the trajectory, as well as the CoM trajectory are taken as inputs to the QP-solver. In contrast, we directly optimize the $z$-component of the robot, thus allowing the optimizer to discover jump-like motions.

The Handle and the Flea by Boston Dynamics show remarkably dynamic motions. Flea is capable of jumping to a height of \SI{10}{\meter} using combustive propulsion \cite{fleaactuator2012}. Handle shows jumping motion while driving forward, as well as traversing rugged terrain. However, methods used to control these systems have not been published.

Indeed, there is a gap with respect to the control of wheel-legged robots. Staged optimization and linearization schemes inherently do not allow for the system to fully exploit its dynamics. Motions such as jumping are hard to execute using these schemes. Jumping in particular is often accomplished using a hand-tuned controller (e.g. Ascento \cite{klemm_ascento:_2019} and AirHopper \cite{tanaka_development_2008}).

Instead of considering the full rigid body dynamics, one can model the system using simpler template models that capture the essence of the robot dynamics.

A popular template model for legged robots is the Spring-Loaded Inverted Pendulum (SLIP)~\cite{poulakakis_spring_2009}. SLIP has been used for high speed running and jumping~\cite{wensing_development_2014}. For wheeled balancing systems, Wheeled Inverted Pendulum models (WIP) have been extensively used as template models~\cite{chan_review_2013}. Template models based on SLIP and WIP can be applied to wheel-legged systems, however they will consider either only the leg or only the wheel, respectively.

In this paper, we propose the simplest template model of wheel-legged robots. 
Our system consists of two sets of wheels connected by a prismatic joint (see \autoref{fig:illustration}). Our main contributions are:

\begin{itemize}
    \item We propose the Variable-Length Wheeled Inverted Pendulum (VL-WIP) template model for wheeled-legged systems.
    \item We implement a motion planner using the VL-WIP model and the direct transcription method which we integrate into a Model Predictive Controller (MPC).
    \item We show that using the VL-WIP model, the robot can execute motions such as jumping that are only achievable using the combined action of wheels and base.
    \item We validate the controller in a closed loop in a physics simulator, and we evaluate the robustness of the MPC with sensor noise and on a rough terrain locomotion task. We also compare these results to a Proportional Derivative (PD) controller on a driving and balancing task.
\end{itemize}

The ability to discover dynamic motion and to exploit the predictive power of the controller stems from the dynamics model we propose.

%
%
%
%
%

\section{Dynamics Model}
\label{sec:dynamics}

\begin{figure}[t]
    \centering
    \begin{tikzpicture}[
        media/.style={font={\footnotesize\sffamily}},
        interface/.style={
            postaction={draw,decorate,decoration={border,angle=-45,
                        amplitude=0.3cm,segment length=2mm}}},
        scale=1.2
        ]
        
        \draw[blue,line width=.5pt,interface](-3,0)--(3,0);
        \draw[dashed,gray](0,0)--(0,3);

        \draw[<->,line width=1pt, shift={(-3cm, 0cm)}] (1,0) node[above]{$x$}-|(0,1) node[left]{$z$};

        \path (0,0)++(80:2.cm)node{$\theta$};
        \draw[->] (0,2.5) arc (90:30:.75cm);

        \path (0,0)++(-1.2, 1)node{$\phi$};
        \draw[->] (-.6,.55) arc (-120:-250:.5cm);

        \draw[gray,dashed](0,1)--(1,.7);
        \path (1.3, .6)node{$x,z$};

        \draw[line width=1pt](0,1cm)circle(.5cm);
        \draw[gray](0,1)--(1.5,2.5);

        \draw[gray, line width=1pt](1.1, 1.9)--(0.9,2.1);
        \draw[gray, line width=1pt](0.9,2.1)--(0.6,1.8);
        \draw[gray, line width=1pt](1.1,1.9)--(0.8,1.6);

        \draw[line width=1pt](1.5,2.5)circle(.1cm);
        \path(1.1, 2.6) node{$m_b$};

        \draw[<->,line width=.5pt,blue,dashed](0.2,0.7)--(1.8, 2.2);
        \path (1.25, 1.25) node{$l$};
    \end{tikzpicture}

    \caption{The proposed Variable-Length Wheeled Inverted Pendulum (VL-WIP) model.}
    \label{fig:dynamics}
\end{figure}
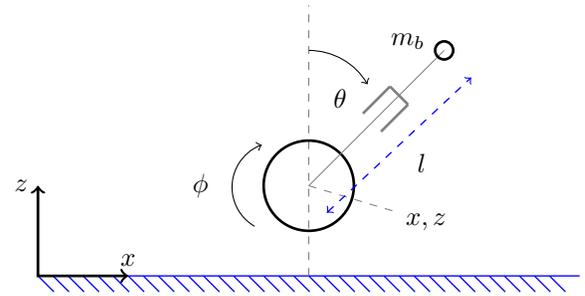

We begin by modeling the system behavior using a template model we call a Variable-Length Wheeled Inverted Pendulum (VL-WIP). Our model is based on the WIP model which is used to control balancing robots~\cite{chan_review_2013}. WIP consists of a wheel and a pole. The pole is modeled as a point mass a fixed distance away from the wheel (see the derivation in \cite{ding_modeling_2012}). We extend this model to include a prismatic joint and a floating base described by a $z$-coordinate as shown in \autoref{fig:dynamics}.
%
%

The dynamics model $\dot{\vx} = \vf(\vx, \vu)$ describes the system behavior in terms of its state $\vx = [\vq^T \; \vqdot^T]^T$ and controls $\vu$, where $\vq$ is the generalized position and $\dot{\vq}$ is the generalized velocity. The generalized position is $\bq = [x \; z \; \phi \; l \; \theta]^T$, where $x, z$ are the coordinates of the center of the wheel, $\phi$ is the wheel's angle along its axis, $l$ is the distance from the center of the wheel to the body point mass and $\theta$ is the rotation of the body from the inertia $z$-axis. 
Controls are $\vu = [\tau \; f]^T$, where $\tau$ is actuation torque of the wheel and $f$ is the linear force in the prismatic joint. 
The explicit Equations of Motion (EoM) for the system under contact can be written as:

\begin{equation}
    \bM(\vq) \vqddot + \bC(\vq, \vqdot) \vqdot + \vg(\vq) = \bS^T \vu + \bJ^T_C \vlam
\end{equation}

where $\bM$ is the mass matrix, $\bC$ is the Coriolis matrix, $\vg$ is the gravity vector, $\bS$ is a selection matrix, $\bJ_C$ is the contact Jacobian and $\vlam$ stands for the contact forces. Note that, this dynamics equation also applies when the robot is in flight and the contact forces vanish.
We compute $\bM$, $\bC$ and $\vg$ using Lagrangian dynamics. For the full derivation, see the Appendix. The results of this derivation are the following matrices:
%
\begin{align}
\label{eq:motion1}
\mathbf{M}(\vq) &=  \begin{bmatrix} m_t & 0 & 0 &  m_b s_\theta &  m_b l c_\theta\\0 &  m_t & 0 &  m_b c_\theta & -  m_b l s_\theta\\0 & 0 &  I_{w} & 0 & 0\\ m_b s_\theta &  m_b c_\theta & 0 &  m_b & 0\\ m_b l c_\theta & -  m_b l s_\theta & 0 & 0 &  m_b l^{2}\end{bmatrix} \\
   \nonumber \\
\label{eq:motion2}
\mathbf{C}(\vq, \vqdot) &=  \begin{bmatrix}0 & 0 & 0 & 2 m_b c_\theta \dot{\theta} & - m_b l s_\theta \dot{\theta}\\0 & 0 & 0 & - 2 m_b s_\theta \dot{\theta} & - m_b l c_\theta \dot{\theta}\\0 & 0 & 0 & 0 & 0\\0 & 0 & 0 & 0 & - m_b l \dot{\theta}\\0 & 0 & 0 & 2 m_b  l  \dot{\theta} & 0\end{bmatrix} \\
  \nonumber  \\
\label{eq:motion3}
\mathbf{G}(\vq) &=  \begin{bmatrix}0 \;\; g m_t \;\; 0 \;\; g m_b c_\theta \;\; - g m_b l s_\theta \end{bmatrix}^T 
\end{align}
%
where $m_t = m_b +  m_w$ is the total mass which sums up the body mass $m_b$ and the wheel mass $m_w$, $I_w$ is the inertia of the wheel, $s_\theta = \sin(\theta)$ and $c_\theta = \cos(\theta)$.

The matrix $\mathbf{S}$ translates controls $\tau$ and $f$ into the generalized coordinates of the system. $f$ directly maps to the prismatic joint. 
The torque $\tau$ applied on the wheel will lead to a counter reaction torque $-\tau$ on the body of the robot.
We write this down as:
%
\begin{equation}
    \mathbf{S} = \begin{bmatrix}0 & 0 & 1 & 0 & -1\\0 & 0 & 0 & 1 & 0\end{bmatrix}
\end{equation}
%
%
%
%
%
Finally, the contact Jacobian relates the velocity of the contact point $\dot{\vr}_C = [\dot{x}_c \; \dot{z}_c]^T$ to the generalized velocities of the robot:
%
\begin{equation}
    \dot{\vr}_C = \bJ_C \vqdot
    = \begin{bmatrix}
        1 & 0 & -R_w & 0 & 0 \\
        0 & 1 & 0 & 0 & 0
    \end{bmatrix}\ \vqdot
\end{equation}
%
where $R_w$ is the radius of the wheel.

Because of our system's simplicity, we have derived an analytic dynamics model. Compared to the full-body dynamics of complex systems, e.g. computed using recursive algorithms, analytic dynamics and their derivatives are faster to evaluate. Using the analytic dynamics allows us to plan motions for both the body and the wheels in a unified way. A unified planning approach allows the solver to plan for motions for which the interaction between wheels and base is key.

\section{Problem Formulation and Control}
\label{sec:control}

\begin{figure}[t]
    \centering
    \includegraphics[width=\columnwidth,trim={0.3cm 0cm 0cm 0cm},clip]{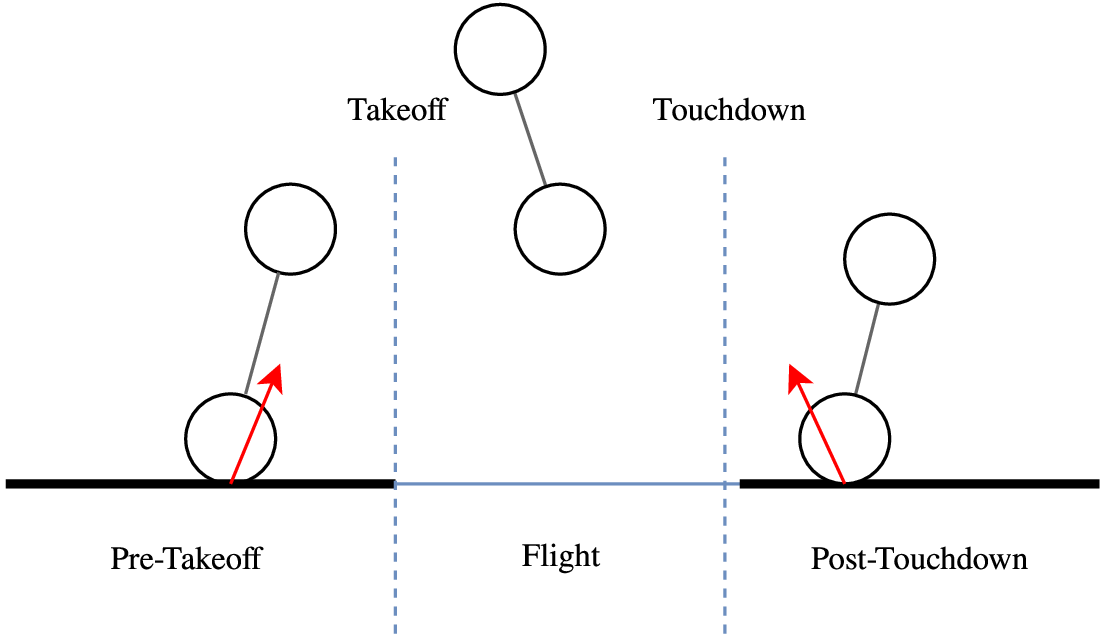}
    \begin{align*}
        \min_{X, U, \Lambda} &
            \ \sum_i W \vu_i^T \vu_i \tag*{(Minimum Energy Cost)} \\
        \text{s.t.}\ &\mathbf{x}_{0} = \mathbf{x}_0^*  \tag*{(Start State)} \\
        &  \mathbf{x}_{N} = \mathbf{x}_N^* \tag*{(Goal State)} \\
        & \mathbf{x}_{t + 1} = f(\mathbf{x}_t, \mathbf{u}_t, \vlam_t),\ t \in [0, N]
            \tag*{(Dynamics)}\\
        &  \mathbf{x}^- \leq \mathbf{x}_{t} \leq \mathbf{x}^+,\ t \in [0, N] \tag*{(State Bounds)} \\
        &  \mathbf{u}^- \leq \mathbf{u}_{t} \leq \mathbf{u}^+,\ t \in [0, N - 1] \tag*{(Control Bounds)} \\
        \text{if}\ &t \in  \;[T_\text{tf}, T_\text{td}]\;\; \text{(Flight Phase)} \\
        &  \bm{\lambda}_\text{t} = \mathbf{0}\  \tag*{(No Ground Force)} \\
        \text{if}\ &t \notin  \;[T_\text{tf}, T_\text{td}]\;\; \text{(Ground Phases)} \\
        & |\bm{\lambda}_{t}^x| \leq \mu \bm{\lambda}_{t}^z, 
        (\mu:\text{friction coefficient}) \tag*{(Friction Cone)}\\
        &  \bm{\lambda}_\text{t}^{z} \geq 0\  \tag*{(Unilateral Force)} \\
        &  \dot{x}_t = R_w\dot{\phi}_t\  \tag*{(No Slip on Ground)} \\
    \end{align*}

    \caption{Problem formulation. On the top we illustrate the three-phases involved in the planning problem. Below is the mathematical formulation of the problem.}
    \label{fig:problem}
\end{figure}

This section looks at how the derived system dynamics can be incorporated into a planning and control pipeline.
For motion planning, we use an optimal control formulation, namely direct transcription \cite{kelly_introduction_2017}. 
To account for model mismatch and perturbations during execution, we use Model Predictive Control \cite{rawlings_model_2017}, which calls the planner iteratively to replan the motion from the current robot state.

\subsection{Planning}
\label{subsec:planning}

The generic planning problem formulation is illustrated in \autoref{fig:problem}.
 We formulate a nonlinear optimization problem (NLP) that finds trajectories $X = \{ \vx_0, \vx_1, \dots, \vx_N\}$, $U = \{ \vu_0, \vu_1, \dots  \vu_{N - 1}\}$, and $\Lambda = \{ \vlam_0, \vlam_1, \dots  \vlam_{N - 1}\}$
 that satisfy the problem constraints and minimize the total energy, weighted by a normalizing term $W$. 
 Here, $\vx$, $\vu$, and $\vlam$ refer to the robot state, controls and contact forces respectively.

Firstly, we specify the duration $T$ of the motion in seconds and discretize it into $N$ knot points. At each knot point the optimizer finds states $\vx$, controls $\vu$, and contact forces $\vlam$ that satisfy the system dynamics $ \mathbf{x}_{t + 1} = f(\mathbf{x}_t, \mathbf{u}_t, \vlam_t)$ derived in \autoref{sec:dynamics}. 

Next we setup task specific constraints.
The start state $\vx_0^*$ and the target state $\vx_N^*$ are specified for knots $\vx_0$ and $\vx_N$. At each knot point, state limits $[\mathbf{x}^-, \mathbf{x}^+]$, and control limits $[\mathbf{u}^-, \mathbf{u}^+]$ are enforced. 
These limits ensure the physical feasibility of the motion generated. 
For different tasks the state and control limits will differ. 
For instance, when jumping over a gap, the state limits for the three phases will indicate where the ``flight`` phase begins and ends.
In general, a jumping motion involves three phases: a ``pre-takeoff" phase, where the robot is on the ground, a ``flight" phase where the robot is in the air, and a ``post-touchdown" phase where the robot lands and re-balances. 
We specify the duration of each phases by setting the time of takeoff $T_\text{tf}$, and the time of touchdown $T_\text{td}$.

Finally, we specify ground contact constraints. 
Contact forces must be zero ($\bm{\lambda}_\text{t} = \mathbf{0}$) while the robot is in the flight phase. 
For the ground phases, they should stay inside the friction cone ($|\bm{\lambda}_{t}^x| \leq \mu \bm{\lambda}_{t}^z$, where $\mu$ is the friction coefficient) and have a positive vertical component ($\bm{\lambda}_\text{t}^{z} \geq 0$). 
For kinematics, we enforce a no-slip constraint while the robot is on the ground: $ \dot{x}_t = R_w\dot{\phi}_t$, where $R_w$ is the wheel radius.
For motions without jumping, we use a similar formulation by removing the flight phase related constraints. 

The planner finds an admissible trajectory for the control task. 
However, for executing it on a real system or in simulation, a controller is needed.

\subsection{Control}
\label{subsec:control}

\begin{figure}
    \centering
    \includegraphics[trim={0cm 0cm 0cm 0cm},clip,width=0.30\textwidth]{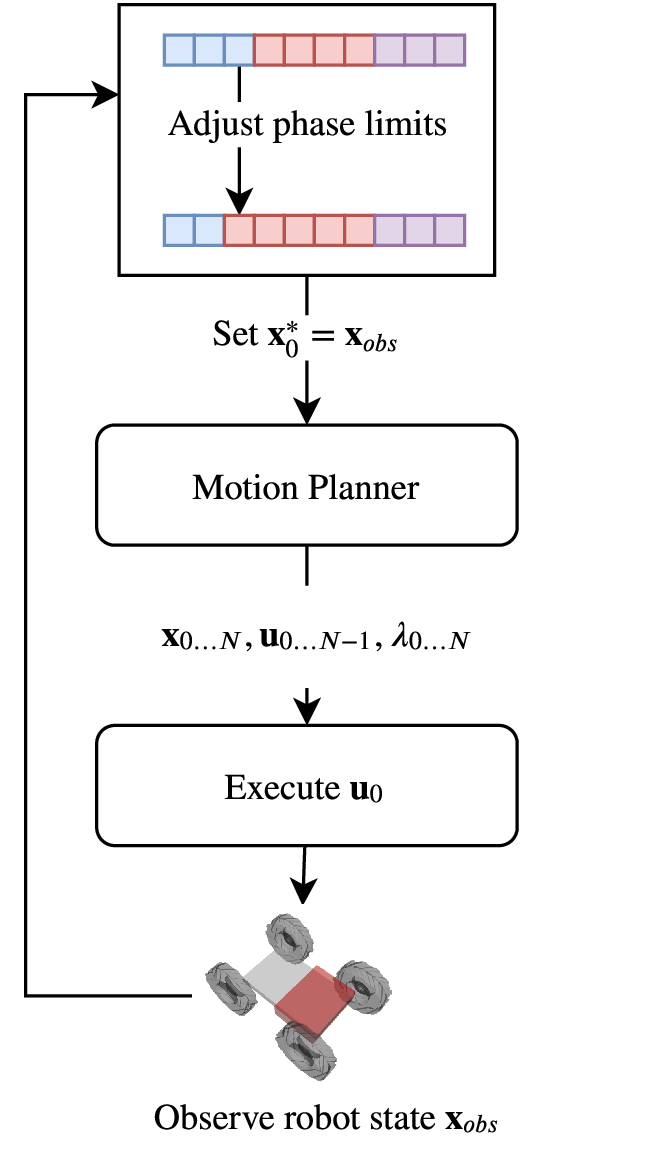}
    \caption{MPC control pipeline. The controller re-plans a motion every timestep by setting the start state $\vx_0^*$ to the current state in simulation $\vx_{obs}$. It then executes the first control $\vu_0$ and updates the phases and limits of the planner.}
    \label{fig:mpc}
\end{figure}

With open-loop control, the planned trajectory cannot be tracked precisely due to tracking error introduced by model mismatch and external perturbations.
Thus, we use Model Predictive Control~~\cite{rawlings_model_2017} to handle the errors.
\autoref{fig:mpc} illustrates the scheme. At each timestep, the planner re-plans the trajectory starting at the observed robot state $\vx_{obs}$ obtained from the simulator. Then, the first action of the plan $\vu_{0}$ is executed. After that, it recedes the time horizon $T$ by the elapsed time.

A natural problem arises with the recession of $T$. Ideally we would keep the number of knot points $N$ constant---this enables saving the optimizer's state and avoids costly re-initialization. The solution requires adjusting the limits of the problem so that each phase always has the same duration. Thus, we maintain the ratio of pre / flight / post phases by shifting the limits of the problem while keeping $N$ the same.


To implement the optimal control problem, we used CasADI~\cite{andersson_casadi_2012}. As the underlying solver, we used KNITRO and selected the Interior/Conjugate-Gradient algorithm that is well-suited for large-scale sparse optimal control problems~\cite{pardalos_knitro_2006}.

\section{Simulation Results}
\label{sec:results}

In order to validate our approach, we investigated several tasks, namely swing-up, balancing and jumping in the physics simulator PyBullet \cite{coumans_pybullet_2016}.
Please find the accompanying video at \url{https://youtu.be/j2sIWL8m2pQ}.

The simulated robot weighs \SI{6}{\kilo\gram} including the wheels and the dimensions of the base are $0.3 \times 0.4 \times 0.1 \si{\meter}$. The radius of the wheels is \SI{0.17}{\meter} and wheels are \SI{0.15}{\meter} wide.


\subsection{Swing-up and Balance}

Firstly, we demonstrate that the robot can switch from a driving mode on four wheels to a balancing mode on two wheels. This demonstrates that the robot can switch between modes without intervention. We used a two-stage controller for this task. Since controlling the robot in driving mode is outside the scope of this paper, we drive forward with a constant acceleration until we reach a velocity of $\dot{x} = 3 \si{\meter\per\second}$. Then we switch to the proposed MPC scheme.

\begin{figure}[h]
    \begin{subfigure}{\columnwidth}
        \centering
        \begin{subfigure}[b]{0.32\textwidth}
            \centering
            \includegraphics[width=\textwidth]{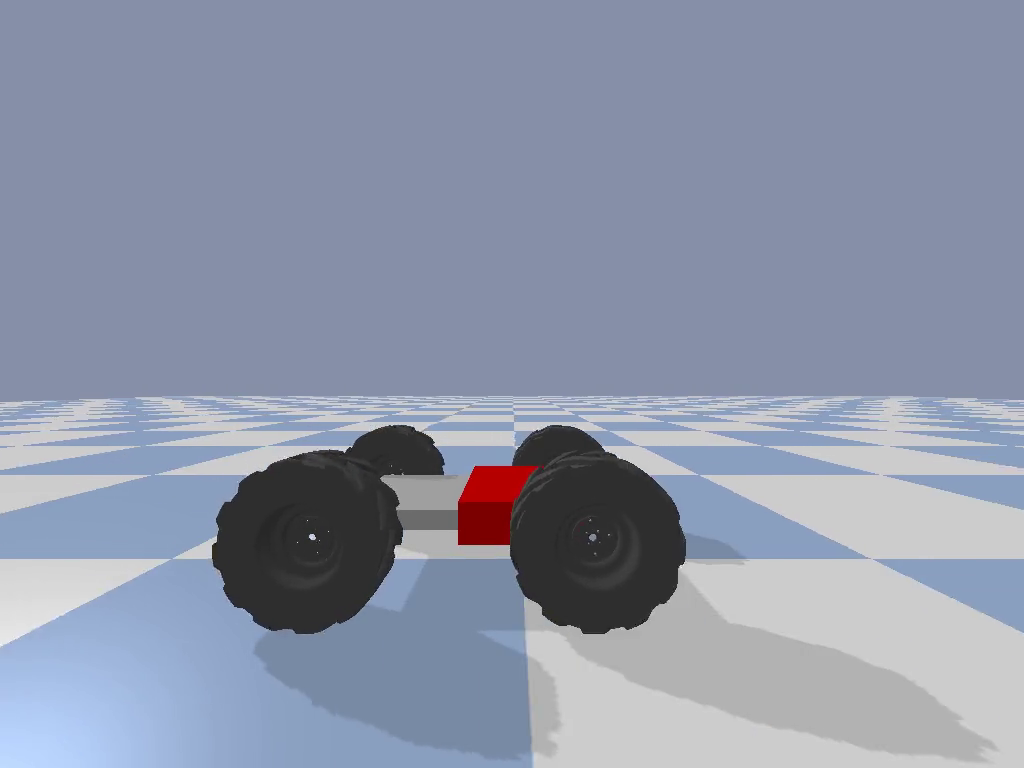}
        \end{subfigure}
        \hfill
        \begin{subfigure}[b]{0.32\textwidth}
            \centering
            \includegraphics[width=\textwidth]{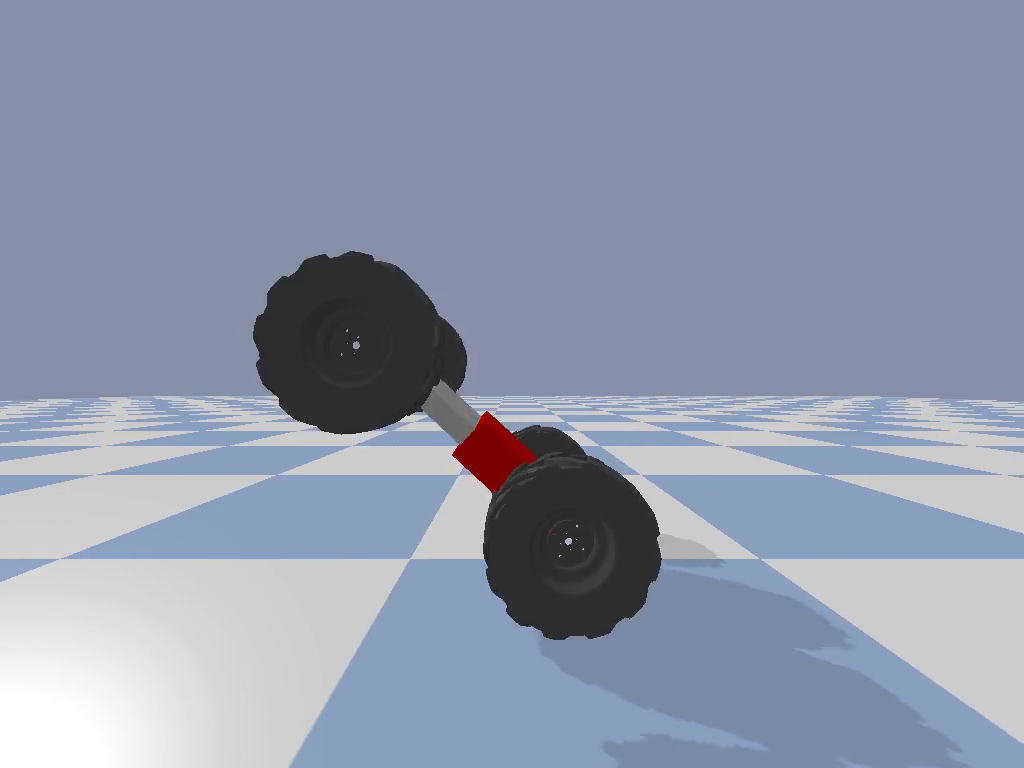}
        \end{subfigure}
        \hfill
        \begin{subfigure}[b]{0.32\textwidth}
            \centering
            \includegraphics[width=\textwidth]{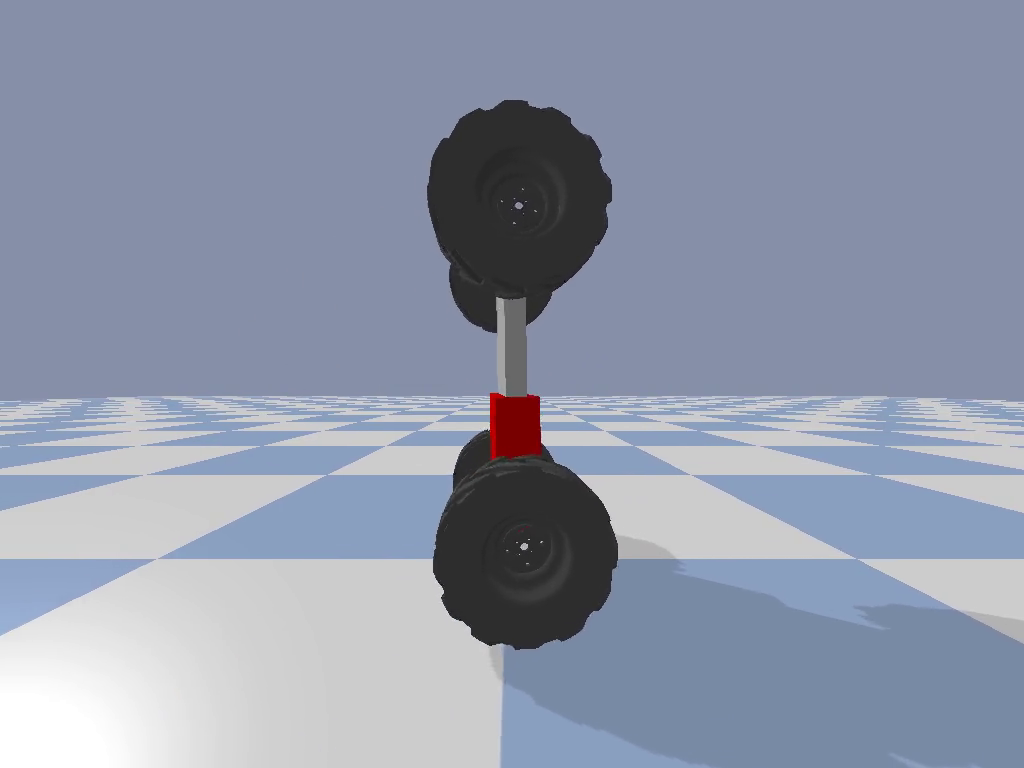}
        \end{subfigure}
    \end{subfigure}
    \caption{Swing-up motion.}
    \label{fig:swingup-full}
\end{figure}

The trajectory duration (and MPC horizon) was $T = 5 \si{\second}$ with $N = 20$ knot points. We set control limits to \SI{\pm 10}{\newton\meter} and \SI{\pm 200}{\newton} for $\tau$ and $f$, respectively.
\autoref{fig:swingup-full} shows the resulting motion. The controller planned a motion where the robot coordinates its prismatic joint and wheel to swing-up. 

\subsection{Driving Upright}
\label{subsec:drive}
Next we demonstrate that, in the upright mode, the robot can balance and drive forward using the proposed approach. 
In this experiment, the goal is to drive the robot forward \SI{1}{\meter} while balancing upright.
We defined the same limits on the states $[\vx^-\; \vx^+]$, and the controls $[\vu^-\; \vu^+]$ for the entire trajectory. We set torque limits for the wheels to \SI{\pm 10}{\newton\meter} and force limits for the prismatic joint to \SI{\pm 200}{\newton}. The total time (MPC horizon) was $T = 3\si{\second}$ and $N = 50$ knot points.

\begin{figure}[h]
    \begin{subfigure}{\columnwidth}
        \centering
        \begin{subfigure}[b]{0.32\textwidth}
            \centering
            \includegraphics[width=\textwidth]{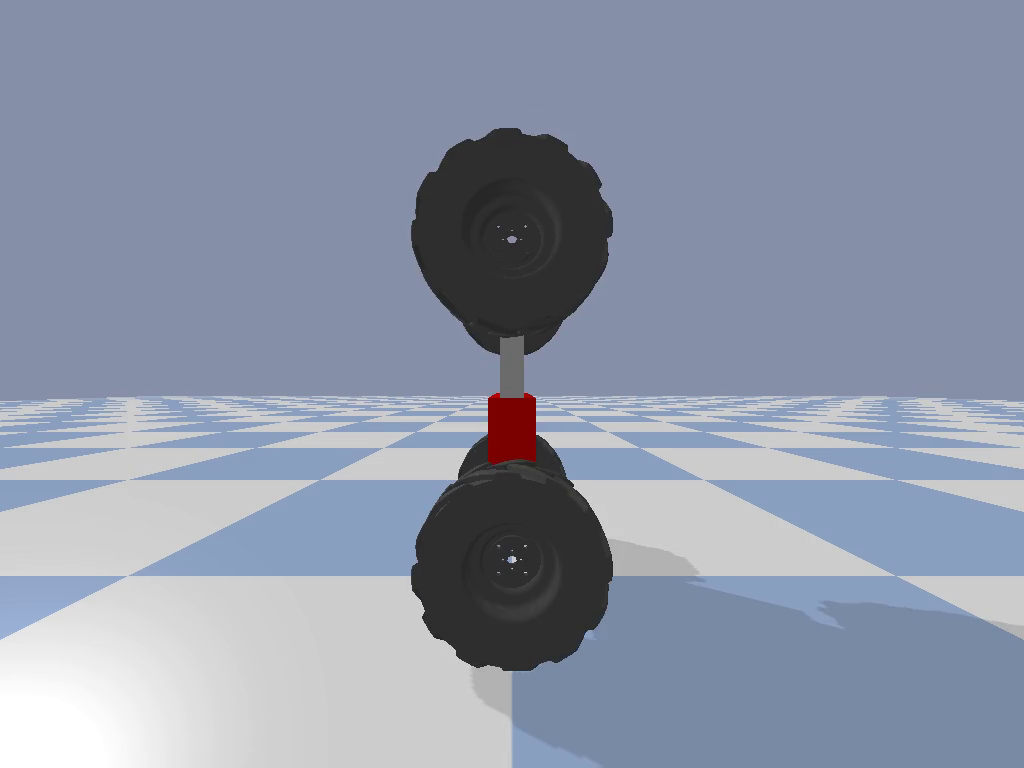}
        \end{subfigure}
        \hfill
        \begin{subfigure}[b]{0.32\textwidth}
            \centering
            \includegraphics[width=\textwidth]{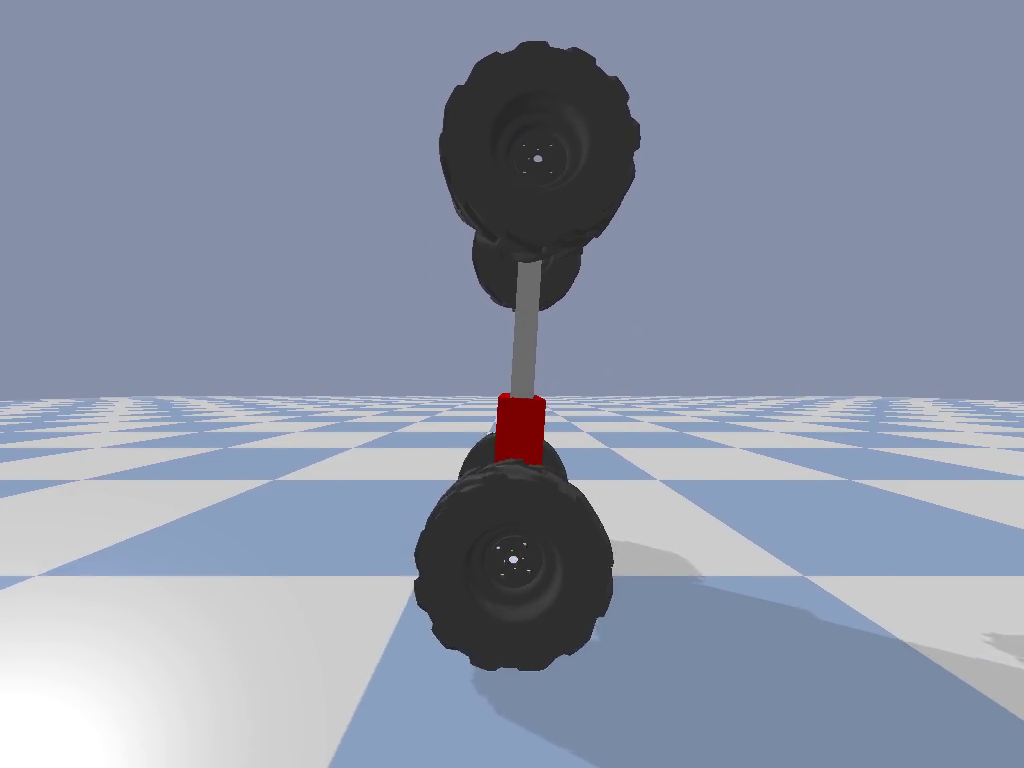}
        \end{subfigure}
        \hfill
        \begin{subfigure}[b]{0.32\textwidth}
            \centering
            \includegraphics[width=\textwidth]{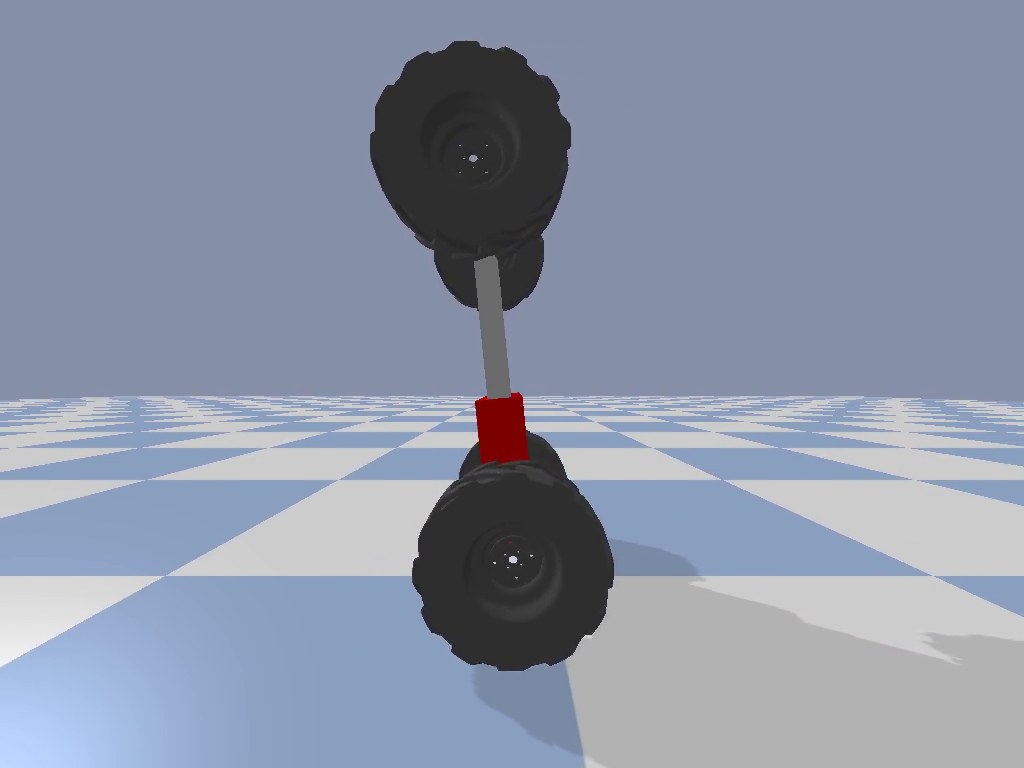}
        \end{subfigure}
    \end{subfigure}
    \caption{Moving forward while balancing.}
    \label{fig:forward-full}
\end{figure}

The resulting motion is shown in \autoref{fig:forward-full}. We noticed that the optimizer first chooses to extend the prismatic joint before moving forward. This behavior can, for instance, be explained by noticing that it is easier to balance a body with a higher center of mass, which is achieved through extending the prismatic joint. Note that we did not in any way encode this behavior---the planner discovered it from first principles of optimization.

\begin{figure*}[t]
    \centering
    \begin{subfigure}[b]{0.16\textwidth}
        \centering
        \includegraphics[width=\textwidth]{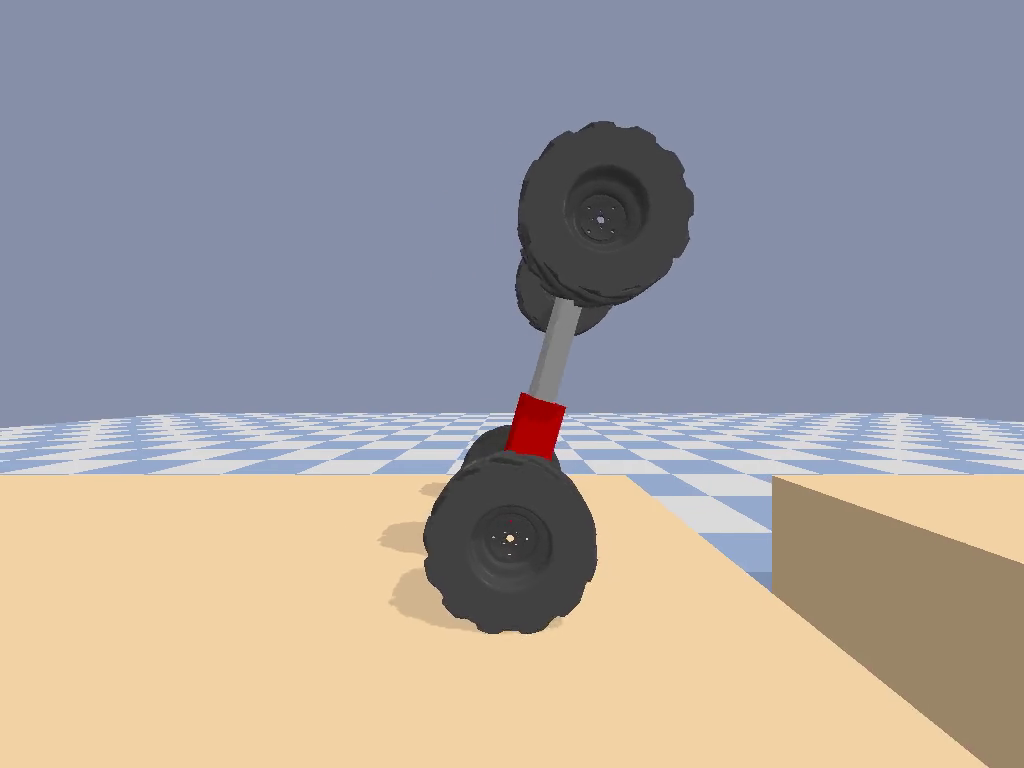}
    \end{subfigure}
    \hfill
    \begin{subfigure}[b]{0.16\textwidth}
        \centering
        \includegraphics[width=\textwidth]{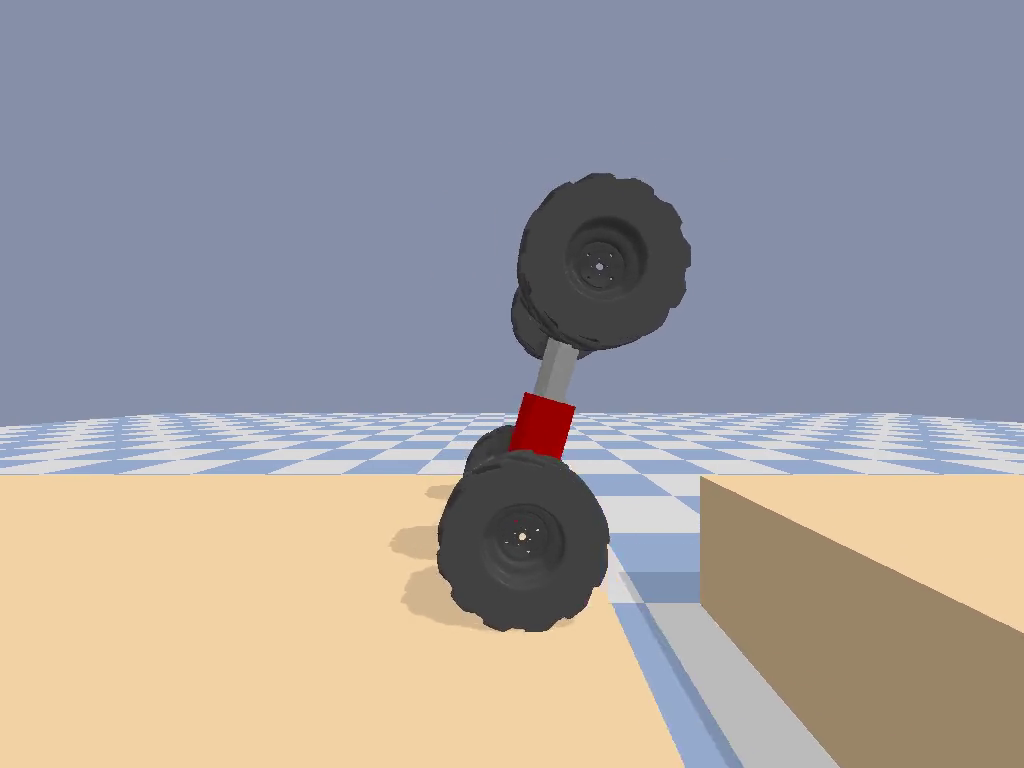}
    \end{subfigure}
    \hfill
    \begin{subfigure}[b]{0.16\textwidth}
        \centering
        \includegraphics[width=\textwidth]{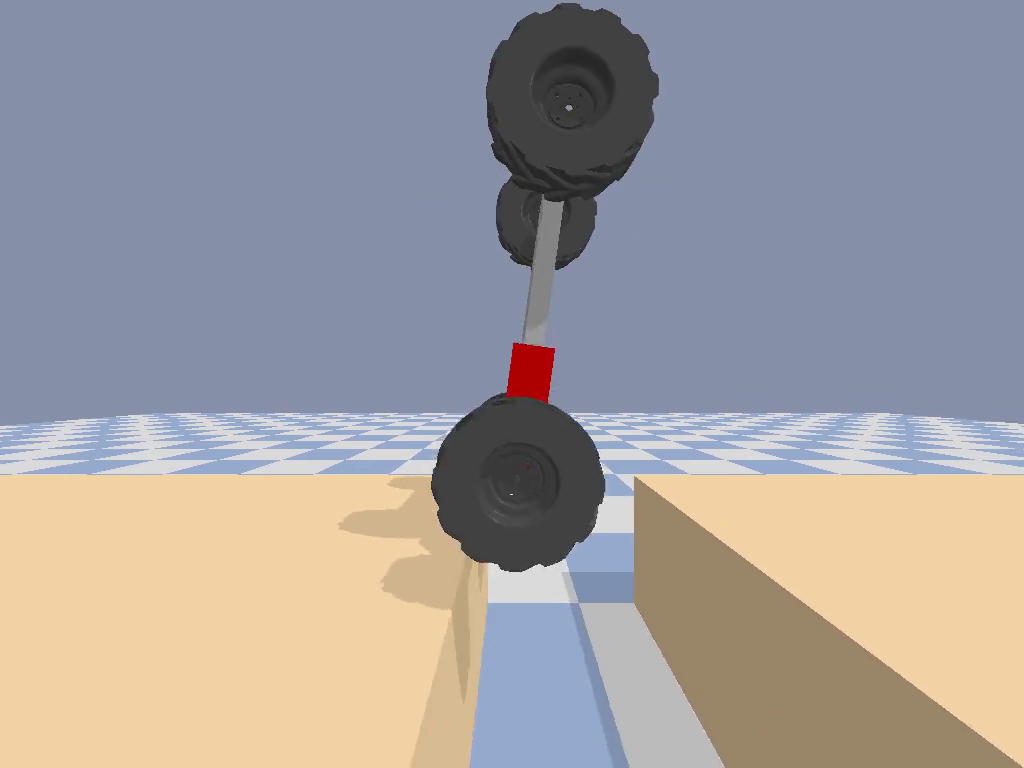}
    \end{subfigure}
    \hfill
    \begin{subfigure}[b]{0.16\textwidth}
        \centering
        \includegraphics[width=\textwidth]{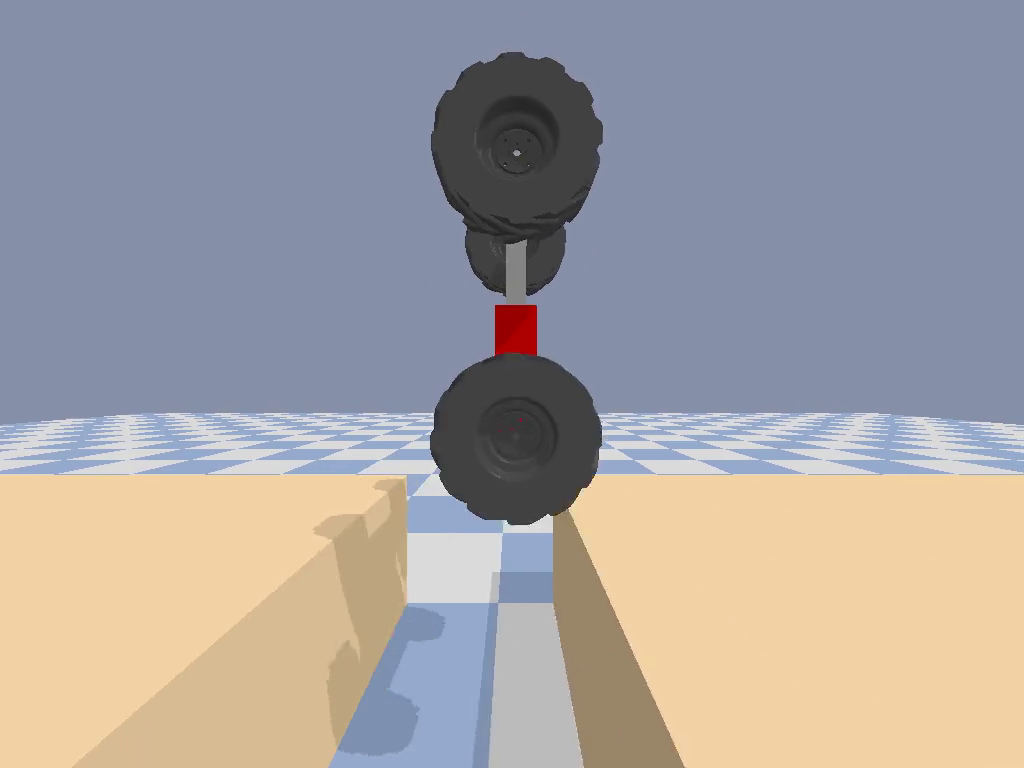}
    \end{subfigure}
    \hfill
    \begin{subfigure}[b]{0.16\textwidth}
        \centering
        \includegraphics[width=\textwidth]{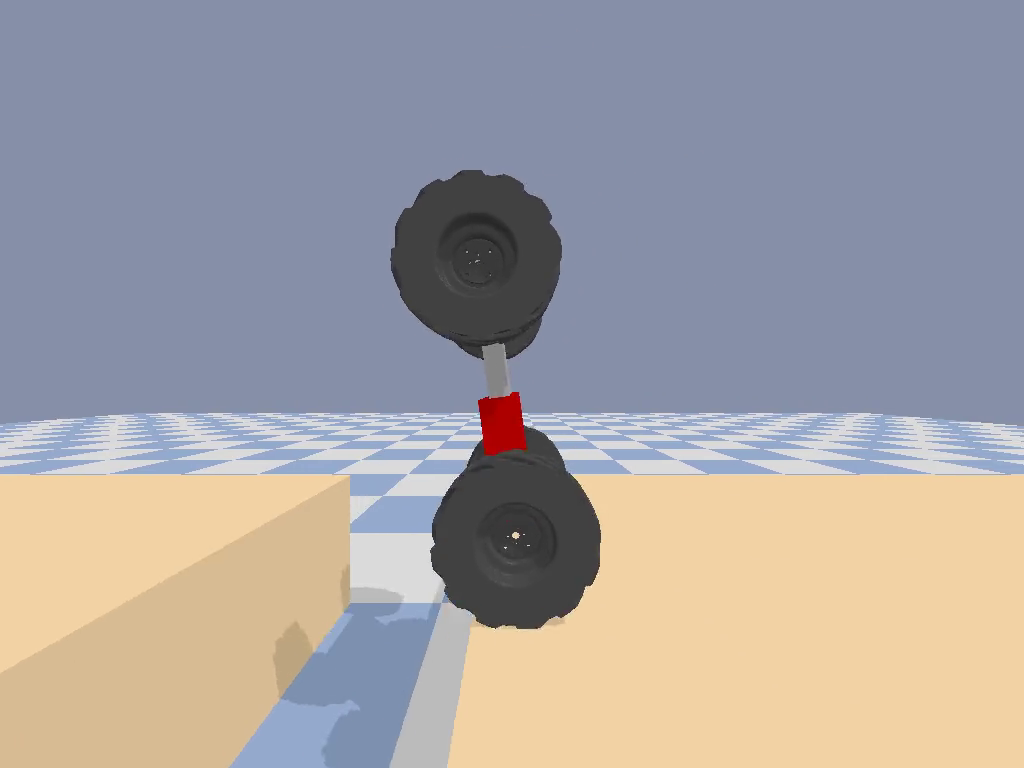}
    \end{subfigure}
    \hfill
    \begin{subfigure}[b]{0.16\textwidth}
        \centering
        \includegraphics[width=\textwidth]{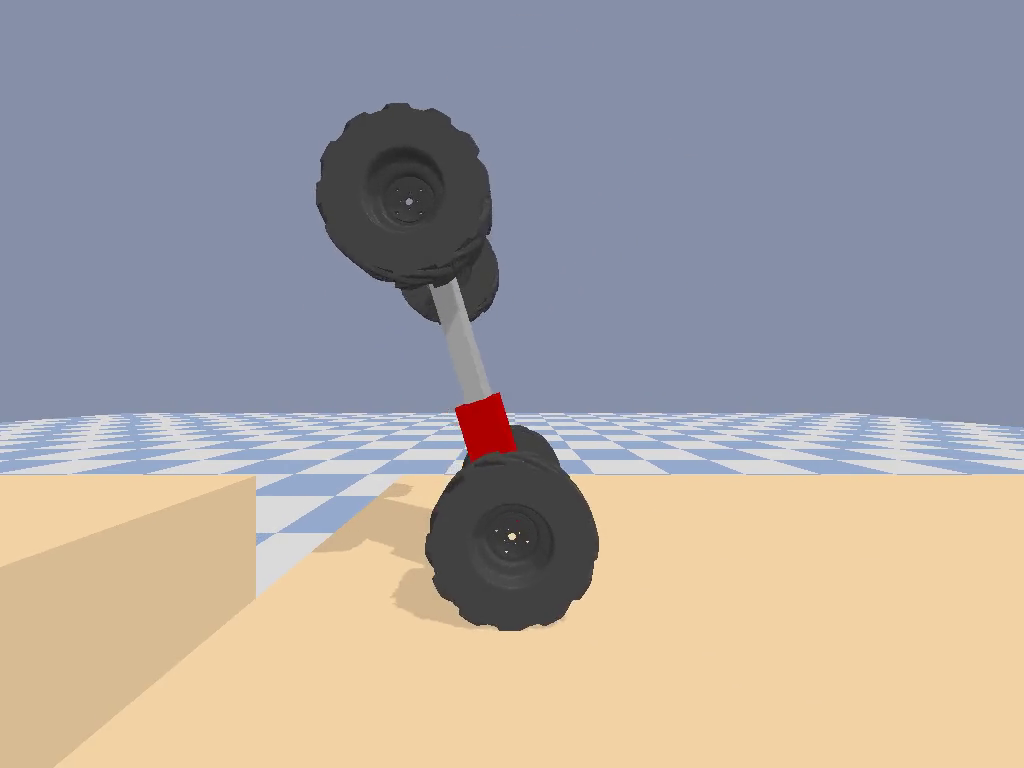}
    \end{subfigure}
    \caption{Jumping over a gap.}
    \label{fig:jump_motion}
\end{figure*}

\begin{figure*}[t]
    \captionsetup{justification=centering}
    \centering
    \includegraphics[width=1\textwidth]{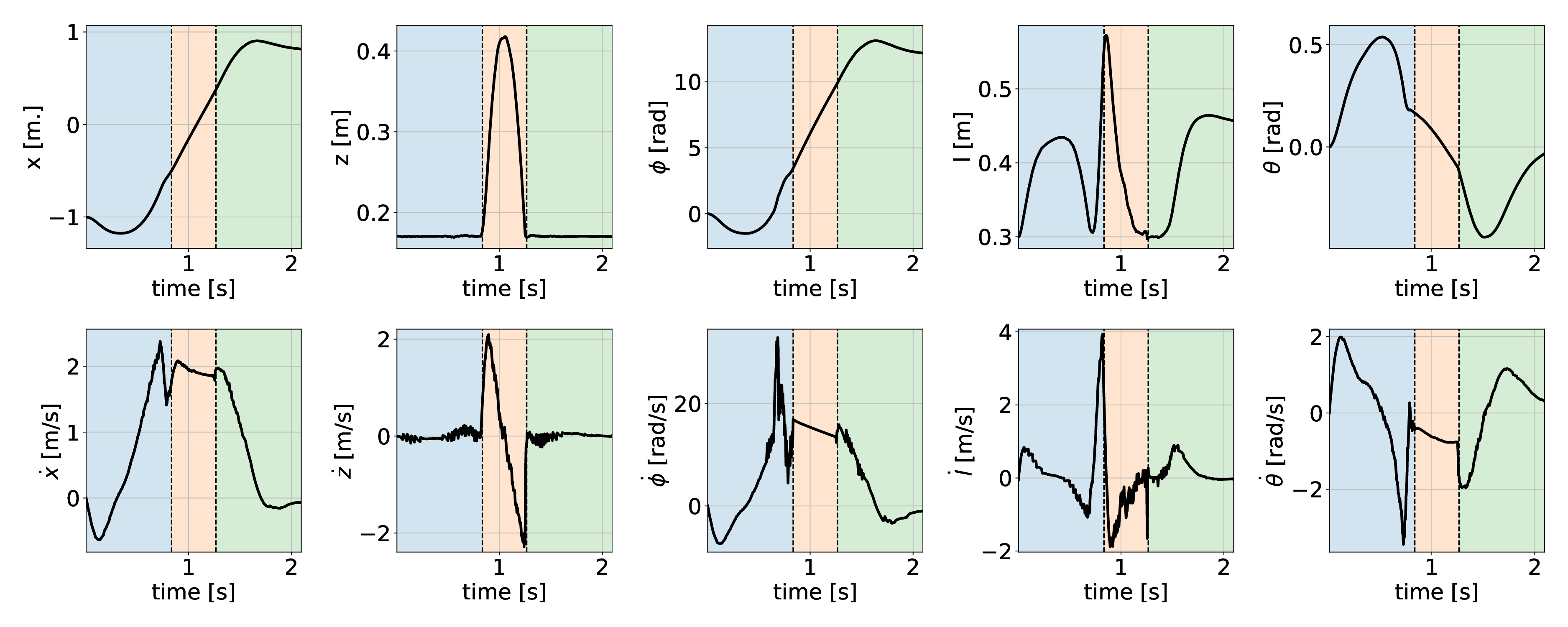}
    \caption{State evolution for the jumping motion in \autoref{fig:jump_motion}. Phases are---``pre-takeoff" in blue, ``flight" in orange, and ``post-touchdown"---in green.}
    \label{fig:jump_trajectory}
\end{figure*}

\subsection{Jumping Over a Gap}

Finally, we consider jumping over a gap, which uses the full multi-phase optimization described in \autoref{sec:control}. This task is more complex since different phases are involved in the motion.

We can entirely describe the jumping motion via state and control limits. In addition to torque limits \SI{\pm 10}{\newton\meter} and force limits \SI{\pm 200}{\newton}, we set the state limits $\vq_{lim}$ for each of the phases. Letting $g^-, g^+$ specify the bounds of the gap in the $x$ direction, we obtain:
%
\begin{align}
    \vq_{lim}^{\textit{pre-tf}} &= \begin{bmatrix}
         -\infty & R_w & -\infty & l^- & -\pi/2 \\
        g^- & R_w & \infty & l^+ & \pi/2
    \end{bmatrix} \\ \nonumber \\
    \vq_{lim}^{flight} &= \begin{bmatrix}
        -\infty & R_w & -\infty & l^- & -\pi/2 \\
        \infty & z^+ & \infty & l^+ & \pi/2
    \end{bmatrix}
    \\ \nonumber \\
    \vq_{lim}^{\textit{post-td}} &= \begin{bmatrix}
        g^+ & R_w & -\infty & l^- & -\pi/2 \\
        \infty & R_w & \infty & l^+ & \pi/2
    \end{bmatrix}
\end{align}
%
where $l^-$ and $l^+$ are the limits of the prismatic joint, $R_w$ is the wheel radius, and $z^+$ limits the jumping height.
In order to set up the length of the trajectory , we ran time optimization within bounds $T^- = 0.3, T^+ = 5$ and knot points $N_{\textit{pre-tf}} = 40, N_{flight} = 20, N_{\textit{post-td}} = 40$. This ensures that the optimizer chooses the appropriate flight duration that is physically feasible within the set time limits. In this sense, $T = 2.3\si{\second}$ was the optimal time and MPC horizon.


The resulting motion is shown in \autoref{fig:jump_motion} and its state and control history is plotted in \autoref{fig:jump_trajectory}. The optimizer first ``contracts" the body of the robot by retracting the prismatic joint before takeoff and extends it just before landing to absorb impact. We can see the preparatory movement in the optimized motion in \autoref{fig:jump_motion} for $x$, $\phi$ and $\theta$.

\section{Robustness}

In the following experiments, we evaluate the robustness of the MPC approach as compared to a Proportional-Derivative (PD) controller with feed-forward torque. 

The PD controller has the following control equation:
%
\begin{align}
    \tau &= \tau_{\text{des}} - K^\theta_p e(\theta) - K^\theta_d e(\dot{\theta}) - K^x_p e(x) - K^x_d e(\dot{x})\\
    f &= f_\text{des} + K^f_p e(f) + K^f_d e(\dot{f})
\end{align}
%
where $e(\cdot)$ is the error between desired and measured state variables, $K^\theta_p = 15$, $K^\theta_d = 2$, $K^f_p = 1,000$, $K^f_d = 100$, $K^x_p=1.25$, $K^x_d=2$ are the PD gains we tuned by hand. We add a term dependent on $x$ in the first equation so that position error is taken into account. Note, the larger PD gains for the prismatic joint---this is due to the scale difference in the length and force needed. We tuned the PD controller so that it achieves similar performance when driving on flat terrain as the MPC controller.





\subsection{Robustness to Sensor Noise}
\begin{figure}
    \centering
    \includegraphics[width=.37\textwidth]{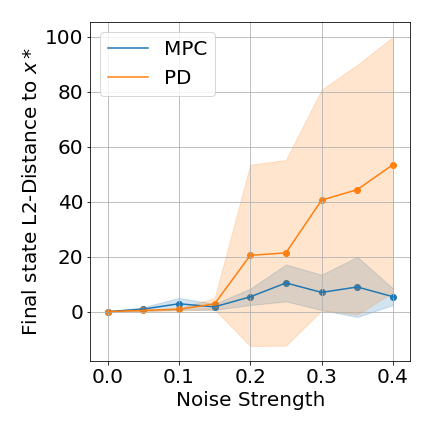}
    \caption{Driving with sensor noise. L2-distance to $\vx^*$ and one standard deviation.}
    \label{fig:robust_noise}
\end{figure}

This experiment has the same setup as \autoref{subsec:drive}. We compare the effects of sensor noise on both the MPC and the PD controllers. We added Gaussian noise to sensor readings at each timestep of the simulation:
%
\begin{equation}
    \mathbf{x}_\text{obs}' = \mathbf{x}_\text{obs} + \mathcal{N}(\mathbf{0}, \sigma \mathbb{I})
\end{equation}
%
where $\sigma$ is the noise standard deviation, which was varied in the $0-0.4$ range and $\mathbb{I}$ is the identity matrix.  We computed the L2-distance from the observed state $\mathbf{x}_\text{obs}$ to the goal state $\mathbf{x}^*$ at the end of the simulation. For this experiment we ran $20$ trials for each value of $\sigma$, computing the means as well as the standard deviations of the L2-distance. The results are in \autoref{fig:robust_noise}.

We notice similar performance for the PD controller at lower noise strengths. However, at $\sigma \geq 0.2$ the PD controller has a significant drop in performance. Additionally, as the noise strength increases, the variability in performance for the PD controller becomes larger. In contrast, the MPC controller shows better mean performance and lower performance variability across $\sigma$ values.

\subsection{Rough Terrain Locomotion}
For this experiment we consider the task of driving forward on two wheels, as outlined in \autoref{subsec:drive}. For the MPC controller we set a target velocity $\dot{x}^* = 1\si{\meter\per\second}$ and horizon $T = 1\si{\second}$. The PD controller has no feed-forward torque reference; it has the same desired velocity target as well as a balance target $\theta^* = 0$. Neither have knowledge of the terrain and the overall duration of each experiment was $T = 5\si{\second}$.

\begin{figure}
    \begin{subfigure}{0.45\textwidth}
        \centering
        \includegraphics[width=0.8\textwidth]{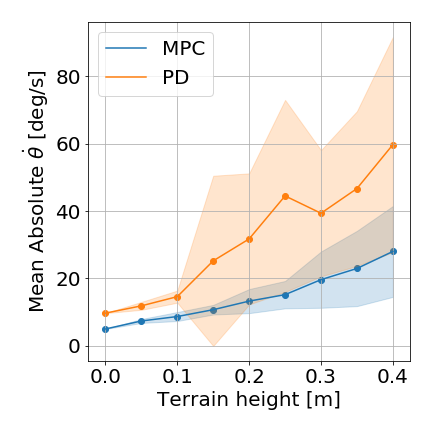}
        \caption{Mean absolute angle velocity $\dot{\theta}$ and one standard deviation.}
        \label{fig:rough_res}
    \end{subfigure}
    \hfill
    \begin{subfigure}{0.45\textwidth}
        \centering
        \includegraphics[width=0.8\textwidth]{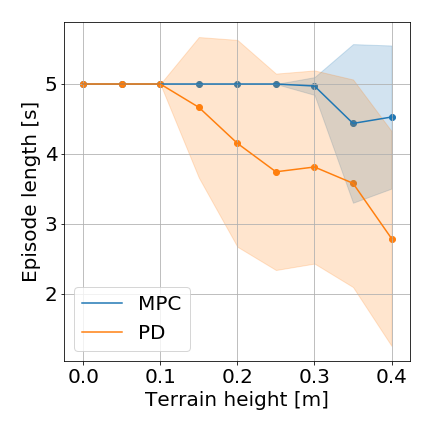}
        \caption{Episode duration and one standard deviation.}
        \label{fig:rough_final_x}
    \end{subfigure}
    \caption{Rough Terrain Locomotion Results}
    \label{fig:rough_loco}
\end{figure}

\begin{figure}[h]
    \captionsetup{justification=centering}
    \begin{subfigure}[t]{0.235\textwidth}
        \centering
        \raisebox{0\height}{\includegraphics[width=1\textwidth,trim={8cm 4cm 10cm 5cm},clip]{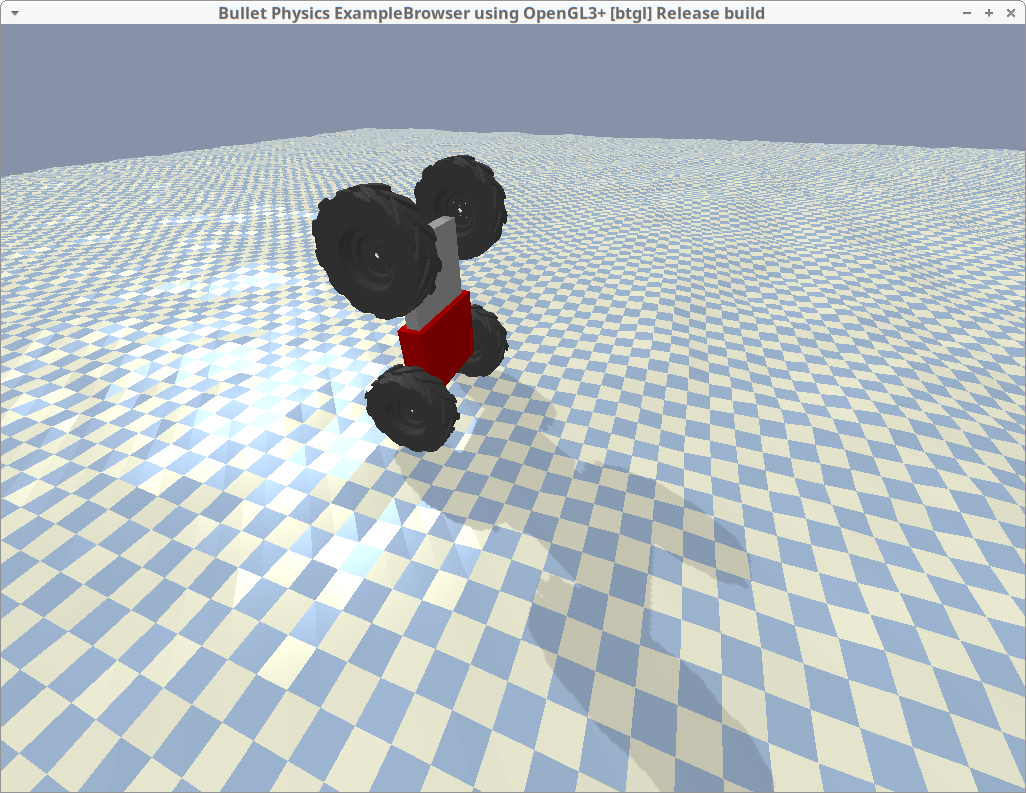}}
        \caption{Rough terrain---height $0.1$m.}
        \label{fig:rough1}
    \end{subfigure}
    \hfill
    \begin{subfigure}[t]{0.235\textwidth}
        \centering
        \raisebox{0\height}{\includegraphics[width=1\textwidth,trim={8cm 4cm 10cm 5cm},clip]{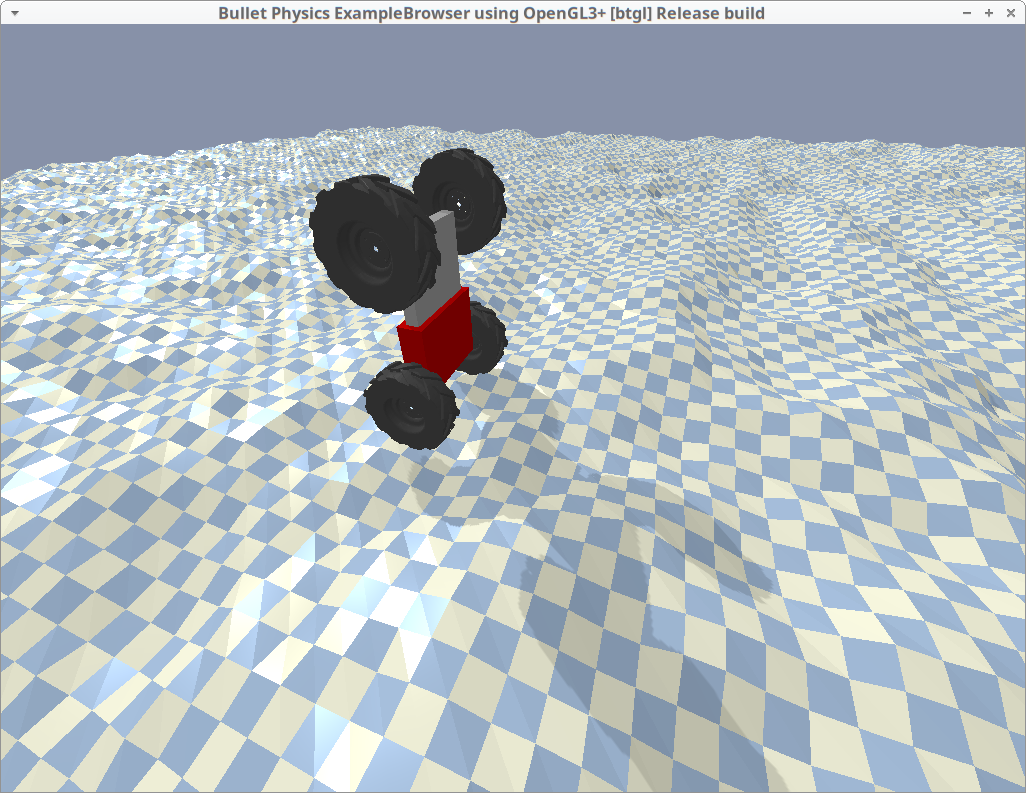}}
        \caption{Rough terrain---height $0.4$m.}
        \label{fig:rough2}
    \end{subfigure}
    \caption{Examples of rough terrain used.}
    \label{fig:rough-terrain-mpc}
\end{figure}

We ran $20$ experiments for varied terrain heights in the range \SIrange{0}{0.4}{\meter}. We generated random terrain using Perlin noise for every experiment while keeping the same random number generator seed between the PD and MPC. Example generated terrain for heights \SI{0.1}{\meter} and \SI{0.4}{\meter} are shown in \autoref{fig:rough1} and \autoref{fig:rough2}, respectively. We computed the mean absolute angle velocity $\dot{\theta}$ for the entire trajectory for each of the terrain heights and its standard deviation. We also show the episode duration as well as one standard deviation. The episode is terminated when the angle of the robot is more than $85$ degrees. The results are in \autoref{fig:rough_res} and \autoref{fig:rough_final_x}, respectively.

For flat terrain and for terrain height up to $\SI{0.1}{\meter}$, MPC and PD show similar results. As the terrain height increases, however, MPC is better at stabilizing the robot with much less variability in performance.



\section{Discussion and Future Work}
The results demonstrated in this paper show that the VL-WIP model, regardless of its simplicity, is sufficient for generating complex dynamic motions. We consider the combined motion planning pipeline developed here as a first step towards more dynamic locomotion of hybrid wheel-legged systems. Future work will apply several connected VL-WIP models, one for each limb, to wheel-legged bipeds and quadrupeds, using the standard motion planning and control pipeline outlined here. This approach is similar to using a SLIP model for each limb of a biped~\cite{wensing_high-speed_2013, wensing_3d-slip_2014,liu_dynamic_2015}. Compared to single rigid-body models, like the one in \cite{winkler_gait_2018}, the stacked VL-WIP model will model leg mass and inertia properties, allowing for more complex in-air maneuvers as well as not requiring inertia compensation.

Furthermore, for all the motions the planner exploited the dynamics of the system without being guided to do so explicitly. Most notably for jumping it discovered the preparatory motion from optimization alone---including contracting and extending the robot before take off, absorbing the impact upon landing, and extending the prismatic joint for easier balancing.


Currently for swing up and jumping we have predefined the switching times. In the future these will be automatically discovered, for instance via phase-based parametrization~\cite{winkler_gait_2018, stouraitis2018dyadic}.

When transferring to a real system, we note that work needs to be done to increase the speed of the computation, for instance by warm-starting the solver using trajectory libraries~\cite{merkt2018leveraging} or using policy learning. 
Our MPC controller ran at $109 \pm 64$ Hz for balancing, $18 \pm 37$ Hz for jumping, and $36 \pm 13$ Hz for swing-up on a Intel Core i9-9980HK at \SI{2.4}{\giga\hertz} with \SI{16}{\giga\byte} of DDR4 RAM at \SI{2667}{\mega\hertz}. We have indeed noticed that the solver did not take an equal number of iterations for all initial conditions and it sometimes did not converge. For this reason the speed was significantly lower on the more complex three-phase optimization for the jumping task. The convergence of the motion planner and the impact of warm starts is outside of the scope of this paper but it presents an exciting direction for future research.




\section*{Acknowledgements}
This research is supported by the EPSRC Centre for Doctoral Training in Robotics and Autonomous Systems (EP/L016834/1), the EPSRC UK RAI Hub in Future AI and Robotics for Space (FAIR-SPACE, ID:EP/R026092/1) and the EU H2020 project Memory of Motion (MEMMO, ID: 780684).
We would like to thank Henrique Ferollho and Matthew Timmons-Brown for their valuable help and feedback,


\bibliographystyle{IEEEtran}
\bibliography{IEEEabrv,IEEEconf,wheels,bib}

\clearpage
\appendix

\label{app:subsec:dynamics}
We use the Lagrangian method to derive the dynamics of the system ~\cite[Chapter~6]{morin_introduction_2008}. 
First, we define the position $x_b, z_b$ of the mass $m_b$ and its velocity $\dot{x}_b, \dot{z}_b$:
%
\begin{align}
    x_b &= x + l\ \sin(\theta) \\
    z_b &= z + l\ \cos(\theta) \nonumber \\
    \dot{x}_b &= \dot{x} + \dot{l}\ \sin(\theta) +
        l\ cos(\theta)\ \dot{\theta} \nonumber \\
    \dot{z}_b &= \dot{z} + \dot{l}\ \cos(\theta) -
        l\ sin(\theta)\ \dot{\theta} \nonumber
\end{align}
%
Lagrange's method states that for a system with total kinetic energy $T$ and potential energy $U$:
\vskip 0.1in
\begin{equation} \label{eq:lagrange}
    \frac{d}{dt} \frac{\partial \mathcal{L}}{\partial \vqdot_i} - \frac{\partial \mathcal{L}}{\partial \vq_i} = \vf_{ext},
\end{equation}
\vskip 0.1in
where $\mathcal{L} = T - U$ is the system's Lagrangian and $\vf_{ext}$ are external forces applied to the system. We now need to compute the system's kinetic and potential energy.

In general, every link will have a rotational and translational kinetic energy component. For the wheel we include a rotational kinetic energy term $I_w \dot{\phi}^2$ where $I_w = m_w R_w^2$ is the moment of inertia of the wheel. Since the point mass has a zero moment of inertia, it only has a translational kinetic energy $m_b \vv_b^T \vv_b$, where $\vv_b$ is the velocity of the point mass.

The only potential energy component is due to gravity $g$ acting on the wheel and the point mass. This leads to:
\begin{align} \label{eq:energy1}
    T &= \frac{1}{2}(I_w \ \dot{\phi}^2 + m_w \dot{x}^2 +
        m_w \dot{z}^2 + m_b \dot{x}_b^2 + m_b \dot{z}_b^2) \\
    U &= m_w g z + m_b g z_b \label{eq:energy2} \\
    \mathcal{L} &= T - U \label{eq:energy3}
\end{align}
Inputting Equations \eqref{eq:energy1}, \eqref{eq:energy2}, and \eqref{eq:energy3} into Equation \eqref{eq:lagrange} leads to a system of five equations, the solution of which is the equations of motion \eqref{eq:motion1}-\eqref{eq:motion3} in \autoref{sec:dynamics}.

\end{document}

%% file: glossary.tex
\newacronym{c-space}{$\mathcal{C}$-space}{configuration space}
\newacronym{com}{CoM}{Centre-of-Mass}
\newacronym{dof}{DoF}{degrees of freedom}
\newacronym{eom}{EoM}{equation of motion}
\newacronym{fk}{FK}{forward kinematics}
\newacronym{giw}{GIW}{Gravito-Inertial Wrench}
\newacronym{ik}{IK}{inverse kinematics}

\newacronym{lq}{LQ}{Linear-Quadratic}
\newacronym{lp}{LP}{linear program}
\newacronym{lping}{LP}{Linear Programming}
\newacronym{nlp}{NLP}{Non-linear Programming}
\newacronym{qp}{QP}{Quadratic Programming}
\newacronym{sdp}{SDP}{Semidefinite Programming}
\newacronym{sqp}{SQP}{Sequential Quadratic Programming}
\newacronym{miqcqp}{MIQCQP}{Mixed-Integer Quadratically Constrained Quadratic Programme}
\newacronym{sip}{SIP}{Semi-Infinite Programming}

\newacronym{lbfgs}{L-BFGS}{Limited-memory BFGS}
\newacronym{bfgs}{BFGS}{Broyden-Fletcher-Goldfarb-Shanno}
\newacronym{prm}{PRM}{Probabilistic Roadmap}
\newacronym{rrt}{RRT}{Rapidly-Exploring Random Tree}
\newacronym{hdrm}{HDRM}{Hierarchical Dynamic Roadmap}
\newacronym{drm}{DRM}{Dynamic Roadmap}
\newacronym{gjk}{GJK}{Gilbert--Johnson--Keerthi}
\newacronym{epa}{EPA}{Expanding Polytope Algorithm}
\newacronym{cmaes}{CMA-ES}{Covariance Matrix Adaptation Evolution Strategy}
\newacronym{pso}{PSO}{Particle Swarm Optimization}
\newacronym{chomp}{CHOMP}{Covariant Hamiltonian Optimization for Motion Planning}
\newacronym{stomp}{STOMP}{Stochastic Trajectory Optimization for Motion Planning}
\newacronym{aico}{AICO}{Approximate Inference COntrol}
\newacronym{riemo}{RieMo}{Riemannian Motion Optimization}
\newacronym{komo}{KOMO}{k-Order Motion Optimization}
\newacronym{pi2}{$\text{PI}^2$}{Policy Improvement with Path Integrals}
\newacronym{trajopt}{TrajOpt}{Trajectory Optimization for Motion Planning}
\newacronym{ddp}{DDP}{Differential Dynamic Programming}
\newacronym{sbp}{SBP}{Sampling-based Planning}
\newacronym{cg}{CG}{Conjugate Gradient}
\newacronym{knn}{KNN}{$k$-Nearest Neighbor}
\newacronym{pca}{PCA}{Principal Component Analysis}
\newacronym{mse}{MSE}{Mean Squared Error}
\newacronym{mae}{MAE}{Mean Absolute Error}
\newacronym{lwpr}{LWPR}{Locally Weighted Projection Regression}
\newacronym{gpr}{GPR}{Gaussian Process Regression}

\newacronym{mpc}{MPC}{Model-Predictive Control}
\newacronym{pbd}{PbD}{Programming by Demonstration}
\newacronym{lfd}{LfD}{Learning from Demonstration}
\newacronym{ioc}{IOC}{Inverse Optimal Control}
\newacronym{irl}{IRL}{Inverse Reinforcement Learning}
\newacronym{oc}{OC}{Optimal Control}
\newacronym{rl}{RL}{Reinforcement Learning}

\newacronym{moe}{MoE}{Mixture-of-Experts}
\newacronym{poe}{PoE}{Product-of-Experts}
\newacronym{gmm}{GMM}{Gaussian Mixture Model}

\newacronym{scd}{SCD}{Smooth Collision Distance}
\newacronym{sop}{SoP}{Sum-of-Penetrations}
\newacronym{cd}{CD}{Collision Distance}
\newacronym{cc}{CC}{Collision Check}

\newacronym{sme}{SME}{Small and Medium Enterprise}
\newacronym{agv}{AGV}{Autonomous Ground Vehicle}
\newacronym{ros}{ROS}{Robot Operating System}
\newacronym{imu}{IMU}{Inertial Measurement Unit}
\newacronym{gps}{GPS}{Global Positioning System}
\newacronym{slam}{SLAM}{Simultaneous Localisation and Mapping}
\newacronym{ukf}{UKF}{Unscented Kalman Filter}
\newacronym{rsi}{RSI}{Repetitive Strain Injury}
\newacronym{oem}{OEM}{Original Equipment Manufacturer}
\newacronym{drc}{DRC}{{DARPA} Robotics Challenge}
\newacronym{eod}{EOD}{Explosive Ordnance Disposal}
\newacronym{idrm}{iDRM}{inverse Dynamic Reachability Map}
\newacronym{mit}{MIT}{Massachusetts Institute of Technology}
\newacronym{exotica}{EXOTica}{Extensible Optimization Toolset}

%% file: iros2020.bbl
\begin{thebibliography}{10}
\providecommand{\url}[1]{#1}
\csname url@rmstyle\endcsname
\providecommand{\newblock}{\relax}
\providecommand{\bibinfo}[2]{#2}
\providecommand\BIBentrySTDinterwordspacing{\spaceskip=0pt\relax}
\providecommand\BIBentryALTinterwordstretchfactor{4}
\providecommand\BIBentryALTinterwordspacing{\spaceskip=\fontdimen2\font plus
\BIBentryALTinterwordstretchfactor\fontdimen3\font minus
  \fontdimen4\font\relax}
\providecommand\BIBforeignlanguage[2]{{%
\expandafter\ifx\csname l@#1\endcsname\relax
\typeout{** WARNING: IEEEtran.bst: No hyphenation pattern has been}%
\typeout{** loaded for the language `#1'. Using the pattern for}%
\typeout{** the default language instead.}%
\else
\language=\csname l@#1\endcsname
\fi
#2}}

\bibitem{klemm_ascento:_2019}
V.~Klemm, A.~Morra, C.~Salzmann, F.~Tschopp, K.~Bodie, L.~Gulich, N.~Kung,
  D.~Mannhart, C.~Pfister, M.~Vierneisel, F.~Weber, R.~Deuber, and R.~Siegwart,
  ``\BIBforeignlanguage{en}{Ascento: {A} {Two}-{Wheeled} {Jumping} {Robot}},''
  in \emph{\BIBforeignlanguage{en}{2019 {International} {Conference} on
  {Robotics} and {Automation} ({ICRA})}}.\hskip 1em plus 0.5em minus
  0.4em\relax Montreal, QC, Canada: IEEE, May 2019, pp. 7515--7521.

\bibitem{bjelonic_rolling_2020}
M.~Bjelonic, P.~K. Sankar, C.~D. Bellicoso, H.~Vallery, and M.~Hutter,
  ``Rolling in the {Deep} -- {Hybrid} {Locomotion} for {Wheeled}-{Legged}
  {Robots} using {Online} {Trajectory} {Optimization},'' \emph{arXiv:1909.07193
  [cs, eess]}, Jan. 2020.

\bibitem{bjelonic_keep_2019}
M.~Bjelonic, C.~D. Bellicoso, Y.~de~Viragh, D.~Sako, F.~D. Tresoldi,
  F.~Jenelten, and M.~Hutter, ``Keep {Rollin}' - {Whole}-{Body} {Motion}
  {Control} and {Planning} for {Wheeled} {Quadrupedal} {Robots},'' \emph{IEEE
  Robotics and Automation Letters}, vol.~4, no.~2, pp. 2116--2123, Apr. 2019.

\bibitem{geilinger_skaterbots:_2018}
M.~Geilinger, R.~Poranne, R.~Desai, B.~Thomaszewski, and S.~Coros,
  ``\BIBforeignlanguage{en}{Skaterbots: optimization-based design and motion
  synthesis for robotic creatures with legs and wheels},''
  \emph{\BIBforeignlanguage{en}{ACM Transactions on Graphics}}, vol.~37, no.~4,
  pp. 1--12, July 2018.

\bibitem{giftthaler_efficient_2017}
M.~Giftthaler, F.~Farshidian, T.~Sandy, L.~Stadelmann, and J.~Buchli,
  ``Efficient kinematic planning for mobile manipulators with non-holonomic
  constraints using optimal control,'' in \emph{2017 {IEEE} {International}
  {Conference} on {Robotics} and {Automation} ({ICRA})}, May 2017, pp.
  3411--3417.

\bibitem{giordano_kinematic_2009}
P.~R. Giordano, M.~Fuchs, A.~Albu-Schaffer, and G.~Hirzinger, ``On the
  kinematic modeling and control of a mobile platform equipped with steering
  wheels and movable legs,'' in \emph{2009 {IEEE} {International} {Conference}
  on {Robotics} and {Automation}}, May 2009, pp. 4080--4087.

\bibitem{nagano_stable_2015}
K.~Nagano and Y.~Fujimoto, ``The stable wheeled locomotion in low speed region
  for a wheel-legged mobile robot,'' in \emph{2015 {IEEE} {International}
  {Conference} on {Mechatronics} ({ICM})}, Mar. 2015, pp. 404--409.

\bibitem{tedrake_underactuated_2018}
\BIBentryALTinterwordspacing
R.~Tedrake, ``\BIBforeignlanguage{en}{Underactuated {Robotics}},'' Feb. 2018.
  [Online]. Available: \url{http://underactuated.mit.edu/}
\BIBentrySTDinterwordspacing

\bibitem{orin_centroidal_2013}
D.~E. Orin, A.~Goswami, and S.-H. Lee, ``\BIBforeignlanguage{en}{Centroidal
  dynamics of a humanoid robot},'' \emph{\BIBforeignlanguage{en}{Autonomous
  Robots}}, vol.~35, no.~2, pp. 161--176, Oct. 2013.

\bibitem{de_viragh_trajectory_2019}
Y.~de~Viragh, M.~Bjelonic, C.~D. Bellicoso, F.~Jenelten, and M.~Hutter,
  ``Trajectory {Optimization} for {Wheeled}-{Legged} {Quadrupedal} {Robots}
  {Using} {Linearized} {ZMP} {Constraints},'' \emph{IEEE Robotics and
  Automation Letters}, vol.~4, no.~2, pp. 1633--1640, Apr. 2019, conference
  Name: IEEE Robotics and Automation Letters.

\bibitem{fleaactuator2012}
A.~Saunders, C.~E. Thorne, and A.~A. Rizzi, ``Environmentally sealed combustion
  powered linear actuator,'' US Patent US9\,238\,967B2, 2012.

\bibitem{tanaka_development_2008}
T.~Tanaka and S.~Hirose, ``Development of leg-wheel hybrid quadruped
  {\textquotedblleft}{AirHopper}{\textquotedblright} design of powerful
  light-weight leg with wheel,'' in \emph{2008 {IEEE}/{RSJ} {International}
  {Conference} on {Intelligent} {Robots} and {Systems}}, Sept. 2008, pp.
  3890--3895.

\bibitem{poulakakis_spring_2009}
I.~Poulakakis and J.~W. Grizzle, ``The {Spring} {Loaded} {Inverted} {Pendulum}
  as the {Hybrid} {Zero} {Dynamics} of an {Asymmetric} {Hopper},'' \emph{IEEE
  Transactions on Automatic Control}, vol.~54, no.~8, pp. 1779--1793, Aug.
  2009.

\bibitem{wensing_development_2014}
P.~M. Wensing and D.~E. Orin, ``Development of high-span running long jumps for
  humanoids,'' in \emph{2014 {IEEE} {International} {Conference} on {Robotics}
  and {Automation} ({ICRA})}, May 2014, pp. 222--227.

\bibitem{chan_review_2013}
R.~P.~M. Chan, K.~A. Stol, and C.~R. Halkyard, ``\BIBforeignlanguage{en}{Review
  of modelling and control of two-wheeled robots},''
  \emph{\BIBforeignlanguage{en}{Annual Reviews in Control}}, vol.~37, no.~1,
  pp. 89--103, Apr. 2013.

\bibitem{ding_modeling_2012}
Y.~Ding, J.~Gafford, and M.~Kunio, ``\BIBforeignlanguage{en}{Modeling,
  {Simulation} and {Fabrication} of a {Balancing} {Robot}},'' p.~22, 2012.

\bibitem{kelly_introduction_2017}
M.~Kelly, ``An {Introduction} to {Trajectory} {Optimization}: {How} to {Do}
  {Your} {Own} {Direct} {Collocation},'' \emph{SIAM Review}, vol.~59, no.~4,
  pp. 849--904, Jan. 2017.

\bibitem{rawlings_model_2017}
J.~B. Rawlings, D.~Q. Mayne, and M.~Diehl, \emph{Model predictive control:
  theory, computation, and design}.\hskip 1em plus 0.5em minus 0.4em\relax Nob
  Hill Publishing Madison, WI, 2017, vol.~2.

\bibitem{andersson_casadi_2012}
J.~Andersson, J.~{\r A}kesson, and M.~Diehl,
  ``\BIBforeignlanguage{en}{{CasADi}: {A} {Symbolic} {Package} for {Automatic}
  {Differentiation} {andOptimal} {Control}},'' in
  \emph{\BIBforeignlanguage{en}{Recent {Advances} in {Algorithmic}
  {Differentiation}}}, ser. Lecture {Notes} in {Computational} {Science} and
  {Engineering}, S.~Forth, P.~Hovland, E.~Phipps, J.~Utke, and A.~Walther,
  Eds.\hskip 1em plus 0.5em minus 0.4em\relax Berlin, Heidelberg: Springer,
  2012, pp. 297--307.

\bibitem{pardalos_knitro_2006}
R.~H. Byrd, J.~Nocedal, and R.~A. Waltz, ``\BIBforeignlanguage{en}{Knitro: {An}
  {Integrated} {Package} for {Nonlinear} {Optimization}},'' in
  \emph{\BIBforeignlanguage{en}{Large-{Scale} {Nonlinear} {Optimization}}},
  P.~Pardalos, G.~Di~Pillo, and M.~Roma, Eds.\hskip 1em plus 0.5em minus
  0.4em\relax Boston, MA: Springer US, 2006, vol.~83, pp. 35--59.

\bibitem{coumans_pybullet_2016}
E.~Coumans and Y.~Bai, ``Pybullet, a python module for physics simulation for
  games, robotics and machine learning,'' \emph{GitHub repository}, 2016.

\bibitem{wensing_high-speed_2013}
P.~M. Wensing and D.~E. Orin, ``High-speed humanoid running through control
  with a {3D}-{SLIP} model,'' in \emph{2013 {IEEE}/{RSJ} {International}
  {Conference} on {Intelligent} {Robots} and {Systems}}, Nov. 2013, pp.
  5134--5140.

\bibitem{wensing_3d-slip_2014}
------, ``{3D}-{SLIP} steering for high-speed humanoid turns,'' in \emph{2014
  {IEEE}/{RSJ} {International} {Conference} on {Intelligent} {Robots} and
  {Systems}}, Sept. 2014, pp. 4008--4013.

\bibitem{liu_dynamic_2015}
Y.~Liu, P.~M. Wensing, D.~E. Orin, and Y.~F. Zheng, ``Dynamic walking in a
  humanoid robot based on a {3D} {Actuated} {Dual}-{SLIP} model,'' in
  \emph{2015 {IEEE} {International} {Conference} on {Robotics} and {Automation}
  ({ICRA})}, May 2015, pp. 5710--5717.

\bibitem{winkler_gait_2018}
A.~W. Winkler, C.~D. Bellicoso, M.~Hutter, and J.~Buchli, ``Gait and
  {Trajectory} {Optimization} for {Legged} {Systems} {Through} {Phase}-{Based}
  {End}-{Effector} {Parameterization},'' \emph{IEEE Robotics and Automation
  Letters}, vol.~3, no.~3, pp. 1560--1567, July 2018.

\bibitem{stouraitis2018dyadic}
T.~Stouraitis, I.~Chatzinikolaidis, M.~Gienger, and S.~Vijayakumar, ``Dyadic
  collaborative manipulation through hybrid trajectory optimization.'' in
  \emph{CoRL}, 2018, pp. 869--878.

\bibitem{merkt2018leveraging}
W.~Merkt, V.~Ivan, and S.~Vijayakumar,
  ``\href{https://doi.org/10.1109/IROS.2018.8593977}{Leveraging Precomputation
  with Problem Encoding for Warm-Starting Trajectory Optimization in Complex
  Environments},'' in \emph{IEEE/RSJ IROS}, Oct 2018, pp. 5877--5884.

\bibitem{morin_introduction_2008}
D.~Morin, \emph{Introduction to classical mechanics: with problems and
  solutions}.\hskip 1em plus 0.5em minus 0.4em\relax Cambridge University
  Press, 2008.

\end{thebibliography}
